\documentclass[conference]{IEEEtran}
\usepackage{titlesec}
\usepackage{adjustbox}
\usepackage{graphicx}
\titleformat{\paragraph}
{\normalfont\normalsize\bfseries}{\theparagraph}{1em}{}
\titlespacing*{\paragraph}
{0pt}{3.25ex plus 1ex minus .2ex}{1.5ex plus .2ex}
\IEEEoverridecommandlockouts
\usepackage{cite}
\usepackage{amsmath,amssymb,amsfonts}
\usepackage{algorithmic}
\usepackage{graphicx}
\usepackage{hyperref}
\usepackage{textcomp}
\usepackage{makecell}
\usepackage{multirow,bigdelim,dcolumn,booktabs}
\usepackage{array}
\usepackage{float}
\usepackage{enumitem} 
\usepackage[caption=false]{subfig}
\usepackage{url}
\usepackage{xurl}
\usepackage{multirow, booktabs}
\usepackage{etoolbox}
\makeatletter
\patchcmd{\@makecaption}
  {\scshape}
  {}
  {}
  {}
\makeatother

\usepackage{xcolor}
\def\BibTeX{{\rm B\kern-.05em{\sc i\kern-.025em b}\kern-.08em
    T\kern-.1667em\lower.7ex\hbox{E}\kern-.125emX}}
\begin{document}

\title{VREN: Volleyball Rally Dataset with Expression Notation Language\\
}

\author{\IEEEauthorblockA{{Haotian Xia$^*$}, {Rhys Tracy$^*$}, {Yun Zhao}, {Erwan Fraisse}, {Yuan-Fang Wang}, {Linda Petzold}}
\textit{Department of Computer Science, University of California, Santa Barbara, CA, USA } 
\\ haotianxia, rhystracy@ucsb.edu
}

\maketitle
\def\thefootnote{* }\footnotetext{ These two authors contributed equally to this paper.}\def\thefootnote{\arabic{footnote}}

\thispagestyle{plain}
\begin{abstract}
This research is intended to accomplish two goals: The first goal is to curate a large and information rich dataset that contains crucial and succinct summaries on the players' actions and positions and the back-and-forth travel patterns of the volleyball in professional and NCAA Div-I indoor volleyball games. While several prior studies have aimed to create similar datasets for other sports (e.g. badminton and soccer),  creating such a dataset for indoor volleyball  is not yet realized. The second goal is to introduce a volleyball descriptive language to fully describe the rally processes in the games and apply the language to our dataset. Based on the curated dataset and our descriptive sports language, we introduce three tasks for automated volleyball action and tactic analysis using our dataset: (1) Volleyball Rally Prediction, aimed at predicting the outcome of a rally and helping players and coaches improve decision-making in practice, (2) Setting Type and Hitting Type Prediction, to help coaches and players prepare more effectively for the game, and (3) Volleyball Tactics and Attacking Zone Statistics, to provide advanced volleyball statistics and help coaches  understand the game and opponent's tactics better.  We  conducted case studies to show how experimental results can provide insights to the volleyball analysis community. Furthermore, experimental evaluation based on real-world data establishes a baseline for future studies and applications of our dataset and language.  This study bridges the gap between the indoor volleyball field and computer science. The dataset is available at: \url{https://github.com/haotianxia/VREN}
\end{abstract}


\section{Introduction}

Volleyball is one of several sports that has seen a significant increase in participation worldwide in recent years. This increase is particularly noticeable in younger age groups, due to the relatively low risk of injury and the teamwork-heavy nature of the sport. Increasing popularity of volleyball at a younger age has led to an increase in the overall level of the sport, which in turn demands more in-depth tactical analysis and advanced strategies. In general, next to player performance, having proper and versatile tactics are the most important factors in winning for high-level games \cite{b1}, \cite{b2}. 

Analytical studies of sports that combine computing assistance with sports have emerged in recent years. These data-driven studies---including team performance prediction and monitoring \cite{b15}, studies on the development of sports \cite{b3}, and analysis of team tactics and player movements---have changed traditional sports paradigms. These applications are not only a valuable aid to the process of the game but also have a significant impact on the training process. Because of its unique nature, baseball is one of the sports where these computer-assisted studies and analyses have been widely implemented \cite{b4}. For example, areas of study include performance analysis on a specific posture \cite{b5}, player performance and lineup predictions\cite{b6}, match outcome predictions \cite{b7},\cite{b9}, tactical preparation aids \cite{b8}, and similar motion retrieval \cite{b10}. Computer-assisted research on other sports that have been recently introduced include evaluating player actions in soccer \cite{b11} and movement pattern recognition in basketball \cite{b12}.
However, to our best knowledge, indoor volleyball has received less attention from sports researchers and there is still much room to explore applications of computer-aided analyses in the sport. For example, the information in a round, such as receiving and passing positions, reveals the team's tactical choices. Being able to present the information of a round in a concise and informative manner is useful for volleyball decision-making and tactical investigation. In order to bridge the gap, we have used our volleyball knowledge (with two authors being former members of NCAA top-10 ranked UCSB Division 1 Men's Volleyball Team) as well as assistance from coaches, trainers, and other volleyball experts (including UCSB Men's Volleyball head coach, Rick McLaughlin, and UNLV Women's Volleyball assistant coach, Cullen Irons) to design a volleyball-specific unified language to describe volleyball rallies. The advantages of a volleyball-specific unified language include allowing people to understand the game as it is played without watching the game videos and providing a mechanism for converting match videos into a computer dataset for on-court and post-play analysis with advanced machine learning aids. As such, we build our dataset by applying our proposed language to manually label players' actions and locations in videos of volleyball matches at both professional and NCAA Division-I levels. More specifically, we record the actions and locations made between opposing teams from serve to score in a rally and collectively, these rally descriptions are used to describe the history of the players and balls and the strategies used in a game. 

Our data collection and analysis efforts have many unique and novel properties.
While there are already many datasets for sports analysis of rugby \cite{b22}, soccer\cite{b11,b23,b25,b28} and basketball\cite{b24,b26,b27}, they cannot be applied to volleyball analysis because of the completely different approaches to the sport. To the best of our knowledge, there are very few datasets for volleyball analysis, limited to a well-known dataset for indoor volleyball \cite{b13} and a dataset for beach volleyball \cite{b21}. Furthermore, current indoor volleyball datasets have two major flaws for team performance and tactical analysis. First, these datasets focus on the recognition of images. This leads to the volleyball dataset labels not being professional enough: they fail to show all volleyball tactical moves and include all technical volleyball statistics, making their dataset difficult to use in high-level analyses. The beach volleyball datasets also have a significant issues because of the difference in rules and the number of players between beach volleyball and indoor volleyball. As a result, the beach volleyball datasets cannot be applied to indoor volleyball; i.e., the analysis is unable to demonstrate all of the strategic play possibilities of indoor volleyball. For example, the setter of indoor volleyball has five setting possibilities (hitters) to choose from, while the beach volleyball setter has only one. Our dataset does not have these limitations and is different from the beach volleyball datasets. Its rich description enables deep learning algorithms to accomplish multiple  tasks to create a connection between computer science and the volleyball game.

In summary, our contributions are mainly three-fold: 
\begin{itemize}
\item We propose a language to represent indoor volleyball from location and action to rally. The language allows a succinct and informative-rich transcoding of video to structured data. It allows people without a volleyball background to better understand the game and helps professional coaches, analysts, and players to retrieve game details for tactical analysis without watching time-consuming match videos. 
\item We introduce a high-level dataset based on our language to advance indoor volleyball analysis and research. 
\item We are the first to introduce three tasks critical to volleyball tactical analysis: Volleyball rally prediction, volleyball statistical analysis, and setting location and hitting type prediction. We propose many machine learning and deep learning algorithms to address these tasks  and analyze how the analysis results can help teams improve decision-making and potentially change the current pattern of indoor volleyball training and competition. The results also validate the usefulness of our descriptive language schemes.

\end{itemize}

\section{related work}

With the increasing popularity of many sports events and the ubiquitous presence of computing devices, there has been notable progress in  building sports datasets and using  automated computer-aided analysis to improve a team's performance and decision making. Currently, the majority of the research is focusing on basketball, soccer, and rugby. However, little research has focused on volleyball. Advanced volleyball tactical exploration remains challenging due to the lack of specialized data and effective analysis techniques. 

In this section, we review related work on sport-specific unified languages and datasets.

\subsection{Sport-Specific Unified Language}

Sport-specific unified languages are used to convert game videos to a simplified data representation so that game information can be retrieved by reviewing the language description without necessity of watching the game. SPADL \cite{b11} and BLSR \cite{b14} propose sport-specific unified languages for soccer and badminton, respectively. They both focus on encoding event data describing single-player actions and concatenating actions as a sequence to represent a whole game. However, neither of these languages can be applied in indoor volleyball because of the sports' different rules and nature. For example, players cannot keep playing if the ball is out of bounds in soccer, and each side has to hit the birdie over the net in one contact in badminton. There is no similar language to describe indoor volleyball events to the best of our knowledge, so we propose a volleyball-specific unified language to show all crucial volleyball events during a game and help people understand the events of a game without watching.

\subsection{Sport Datasets}

Various datasets for sports analysis have been introduced to empower different ways to collect matches' information in a wide range of sports, such as basketball \cite{b24,b26,b27},
rugby \cite{b22}, badminton \cite{b14}, baseball \cite{b6,b7,b8,b9}, and soccer\cite{b11,b23,b25,b28}. These datasets are each created with different variables based on their sport's unique characteristics and patterns to enable different analysis tasks. Generally, sports analysis tasks can be divided into two areas: match outcome predictions (e.g., \cite{b7,b9,b14,b22,b23,b24,b25,b26,b27,b28}) and player performance analysis (e.g., \cite{b6,b8,b11,b14}). Two of our tasks similarly focus on outcome predictions. However, though the datasets and models proposed above for outcome prediction may perform well in their own sports, they cannot be applied to volleyball prediction because of the sport's different rules and play styles. 

Moreover, to the best of our knowledge, the current primary dataset for volleyball \cite{b13,b21} cannot meet our needs. Specifically, the main indoor volleyball dataset's \cite{b13} labels are not designed to capture all relevant volleyball actions on the court. For example, the dataset does not have variables for the middle blockers' spike and only distinguishes between the left and right side attacks, but not between the front and back row attacks. It is worth noting that the impact of the middle blockers' spike and the attack from the back row or front row are the most important factors in scoring. The beach volleyball dataset \cite{b21} also cannot be applied to our tasks since in beach volleyball there are only two players on each side--so there is only one other player to set the ball to--and there is no distinction between front and back rows. In indoor volleyball, however, there are multiple players that the setter can target, and there is a distinction between front and back rows. Therefore, any beach volleyball datasets lack the characteristic information of indoor volleyball, making them impossible to be utilized for indoor volleyball in general, let alone to fulfill our tasks. Therefore, we curated a dataset that contains appropriate and relevant indoor volleyball features. Furthermore, we show the utility and completeness of our dataset by applying the dataset to three volleyball tasks. Those tasks could potentially bring a new direction, especially in decision making, to the development of volleyball.

\section{Dataset}

\subsection{Necessary Volleyball Knowledge}
In order to understand our data set and descriptive language, it is important to first learn some general rules and terms of volleyball. A volleyball match is split into up to five sets. Each set is played to 25 (sets one to four) or 15 points (fifth set) and a team needs to win by 2 points to win a set. Most NCAA and professional volleyball games are played best of five, that is the first team to win three sets out of five wins the match, but volleyball can also be played as best of three. Only the fifth set in best of five format or the third set in best of three is played to 15 points if it occurs.

Each volleyball rally starts with a serve and ends with a point, and each rally involves consecutive plays where each team can make up to three contacts to send the ball back over the net. The first contact of the ball by a team is considered a pass, but it is also called a dig if the previous contact was a spike by the opposing team and not a serve or a free ball (sent over the net in any way but a one-hand overhead hit). The second contact is considered a set and  is usually done with two hands overhead,  but the second contact is not a set if a player decides to jump and spike the ball. The third contact is considered a hit or a spike unless it is contacted with an underhand "bump" (then it is considered a free ball). Hitters have several strategies they can use when hitting a ball: a hit is when a hitter makes a powerful spike, a roll shot is when the hitter lightly spins the ball in an upward arc to target an area with no defenders, a tip is when the hitter lightly taps the ball with slow speed and no spin using their fingertips, an off-speed hit---usually used when a hitter is in an uncomfortable position---is a hit but with much less speed and spin than normal, and a dump is when the setter throws the ball across the net with power using one hand during the second contact. 

A volleyball team has six players on the court at any point in time: three in the front row (closer to the net) and three in the back row (further from the net) separated by the ten foot line (or the three meter line) which is ten feet from the net on both sides. Although players have to rotate clockwise with each point they win when receiving a serve, they are allowed to swap positions after the ball is served, so long as the back row players stay in the back row. For nearly all strategies and levels of play today, a team includes six set positions: a front row and back row outside, a front row and back row middle blocker (middle), an opposite (oppo), and a setter. The two outsides, the two middles, and the setter and opposite are all situated directly opposite from each other in the rotational lineup. So one of the two in the pair is always in the front row, and the other is always in the back row.

The outsides' positions are on the court's left front and middle back when looking at the net. The middle's positions are in the middle front and left back of the court. Most volleyball rules allow a special defensive (passing/digging) player called the libero to replace a back row player at any time before a rally starts without using a substitution. The libero usually replaces the back row middle blocker as they are usually the worst at back-row defense. The opposite and setter are positioned on the right front and back of the court. There are many terms for the range of sets a setter can set to each player, but they can---for the most part---be generalized to the specific positions they are set to. A set to the front row outside can be considered an outside set. A set to the front row middle can be considered a quick set. A set to the opposite in the front row can be considered an opposite or oppo set. A set to the back row outside is usually called a bic. A set to the opposite in the back row is usually called a d-ball.

When a player is jumping to hit a ball, the opposing team is allowed to have any front row players jump and reach their arms over the net to try to block the ball from crossing over the net. If a blocker touches a ball, it is not considered as one of that team's three contacts. A rally ends when the ball touches the floor, a wall, or one of the antennas protruding from the top of the net on either side or if an error occurs (a player touches the net, a team contacts the ball 4 times before sending it over the net, a player holds the ball, the libero hits the ball over the net, and several more). The overall rules and strategies for volleyball are far more complex than this, but this should be sufficient for a general introduction to the sport to enable readers to understand our data representation at a high level.

\subsection{VREN: Volleyball Rally Expression Notation}
The important terms and features that people use in describing players, locations, actions, and interactions in different sports are often suggested by experts. Using these sports terms and vernaculars enables people to easily communicate specific movements and tactics. Data are usually composed of these  sport-specific terms in sports analysis. By applying the corresponding sports terms in evaluating the matched videos, the experts convert the video content into a uniform data format through manual annotation. For example, SPADL \cite{b11} is a language to integrate event stream data formats to improve data analysis performance in soccer analysis. BLSR \cite{b14} is another language that has been proposed to help improve badminton data analysis, specifically for singleton matches. 

However, as mentioned before, no similar data descriptor exists for indoor volleyball sports analysis. Additionally, no previous formats and descriptors can be easily transcoded from other sports to indoor volleyball. Since SPADL was designed for soccer's unique court setup, movement patterns, and unlimited team and individual possession time and BLSR was designed for badminton's single contact per side and unique types of shots, these previous languages cannot be applied to indoor volleyball where teams get a maximum of three contacts per side \cite{b20} and use very unique strategies.

To address the problem of finding a suitable event descriptor scheme in volleyball and communicating the volleyball characteristics of multiple ball touches, we propose VREN to formalize event stream data after consulting with many volleyball experts.

VREN is a designed language used to describe the volleyball rally process. This standard language makes it possible to succinctly describe all on-court situations, including player movements, positions, and coordination between players, without watching the game's video. The basic scoring unit in volleyball is a rally. Each set contains a number of rallies, and each rally consists of
several movements by two teams. Our description of a rally is a sentence comprising words representing different actions and situations as well as the location on the court where each action takes place. In more detail, a rally, $R_1$, can be interpreted as a sequence of rounds {$r_1^{(R_1)}, r_2^{(R_1)}, r_3^{(R_1)}, r_4^{(R_1)}, r_5^{(R_1)}, r_6^{(R_1)}, ..., r_n^{(R_1)}$}, where n is the total number of rounds of rally $R_1$. Each rally has overall rally-level information, $I^{R_1}$, which includes three variables: 1) winning reason, 2) losing reason, and 3) which team wins the rally. A volleyball match can then be represented by a tuple of rallies {$R_1, R_2, R_3, R_4, R_5, ..., R_N$}, where N is the total number of rallies in the game, equal to the total number of points scored by both teams. Experts suggest that differentiating between the sets of the match is not likely to yield useful information for the purposes of VREN, so we do not include set information in this description.

Each round of each rally, $r_n^{(R_N)}$, is composed of different volleyball experts' suggested variables, such as: 
\begin{enumerate}
  \item round: the number of the current ball round in the rally.
  \item team: which team has the ball or is receiving the serve.
  \item various locations: locations on the court where each contact occurs, where the ball trajectory is heading, and where a digger/passer moves from before digging/passing
  \item pass{\_}rating: rating of the pass.
  \item set{\_}type: rating of the set from the setter.
  \item hit{\_}type: type of attack.
  \item num{\_}blockers: the number of blockers.
  \item block{\_}touch: if blockers touch the ball.
  \item serve{\_}type: type of serve.
\end{enumerate}

\begin{figure}[htbp]
\centerline{\includegraphics[width=7cm, height=3.5cm]{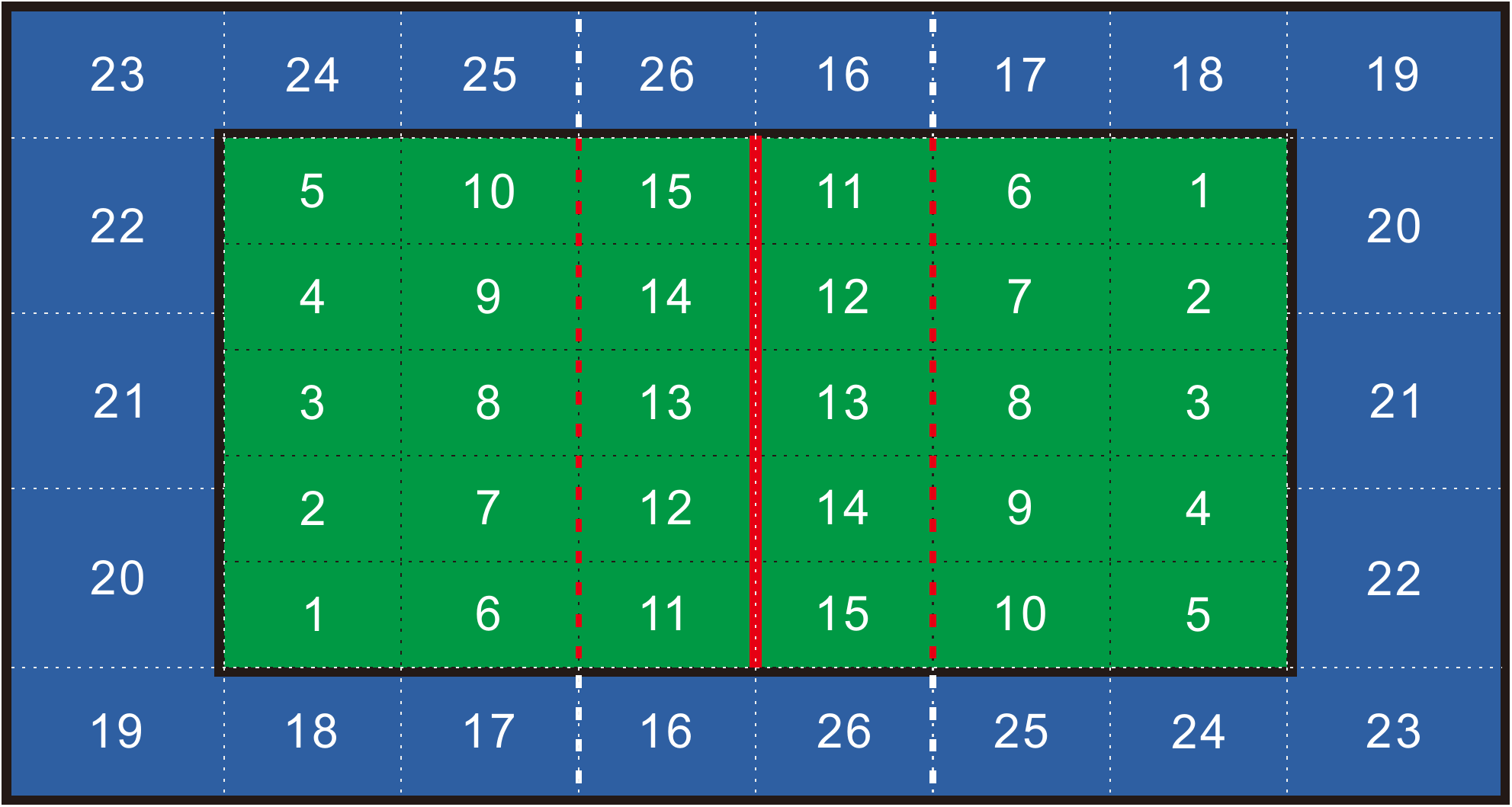}}
\caption{The grid system we propose to represent positions on the volleyball court. The thick solid red line represents the net. The court is symmetrically and uniformly divided into 26 areas on both sides of the red line (the net).}
\label{fig1}
\end{figure}

\begin{figure*}[htbp]
\centering
\includegraphics[width=1\textwidth,height=3.5cm]{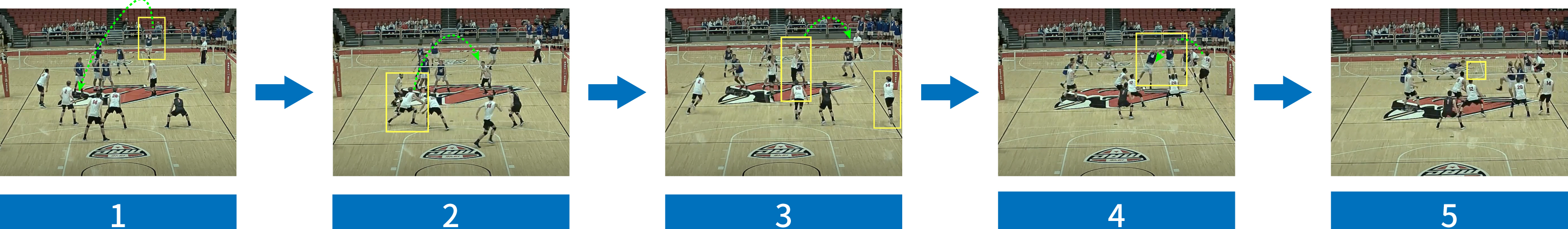}
\caption{Video screenshots showing how we identify our key variables to format individual ball rounds in the VREN language representation. The green dash represent the trajectory and direction of the ball. Based on the ball trajectory and our location information and volleyball terms, we came up with VREN.}
\label{fig2}
\end{figure*}
 
To understand many of these elements require us to explain how we partition the court into grids and how pass rating, hit type, and serve type are recorded. In sports, the choice of tactics and movements is usually determined by the relative position of the ball and the person on the court. Therefore, positioning on the court is important information for sports games. Based on the opinions of volleyball experts, we propose a grid system to divide each side of the volleyball court, including the out of bounds area, into 26 smaller zones, as shown in Fïigure~\ref{fig1}. According to the rule \cite{b20} that the volleyball 3-meter line is a marker to distinguish the front and back rows, we set the areas 16, 11, 12, 13, 14, 15, 26 as the front row, and the rest of the areas are considered as the back row. The grid system symmetrically and uniformly divides the volleyball court on both sides of the net. The areas marked 16 to 26 represent the areas outside the court (out of bounds). The reason for labeling the areas outside the court is based on the rule \cite{b20} that a volleyball can still be contacted outside the court for a rally to continue. Through our proposed grid system, we are able to measure the position of the player at the time of the hit, the quality of the pass, and the tactics of the hit. This grid system helps convert the video information into our proposed VREN language. For example, when describing a player hitting in zone 16 or 11 in our language, it means the opposite made the play in the original video.

After consulting with relevant volleyball coaching experts and according to volleyball rules, we define and explain some of the variables presented above combined with our grid system as follows.

\begin{enumerate}
  \item pass{\_}rating:
  \begin{enumerate} 
    \item in system: pass landing in area 11, 12, 13.
    \item out of system: pass landing in other areas.
    \item overpass: ball passed over net on the first contact.
    \end{enumerate}
  \item hit{\_}type:
   \begin{enumerate} 
    \item hit: hitters make a powerful spike.
    \item off{\_}speed: hitters make a low-speed spike.
    \item roll{\_}shot: hitters make a roll shot.
    \item tip:  hitters tip the ball.
    \item free{\_}ball: hitters do not make a spike and instead pass the ball over the net.
    \item dump: setter dumps the ball.
    \end{enumerate}
  \item serve{\_}type:
  \begin{enumerate} 
    \item float: servers use float serve step approach, served ball has minimal spin and slow speed.
    \item jump: servers use jump serve step approach, serve has a high speed and spin.
    \item hybrid: servers use a float serve approach to make a jump serve or use jump serve step approach to make a float serve.
    \end{enumerate}
\end{enumerate}

Figure~\ref{fig2} gives a round's example of how we used VREN to create a dataset by converting the information from game videos into our language representation. First, we see the defender for team A (white jerseys, near court) who ends up passing the ball standing at location 9 while the server on team B (blue jerseys, far court) tosses the ball up and executes a jump serve (as shown in image 1). Next, we see this defender receiving the ball at location 9 (as shown in image 2) and passing the ball to location 13 (as shown in image 3). Next, the setter moves to location 13, where he then sets a d-ball (as shown in image 3). We then see the opposite hitter on team A jump and hit the d-ball in location 6 while facing two blockers who do not touch the ball (as shown in image 4); the ball then heads directly for location 8 (as shown in image 5). As a result, team A won the point with a kill.

\subsection{Data Collection}
Our experimental data was obtained from real-world men's volleyball tournaments at the national-team and NCAA levels. The game videos were manually annotated by volleyball coaches and experts using our VREN language. We selected games between different teams for our source data in order to ensure statistical diversity because focusing on two teams only can inadvertently cause bias in the use of personnel and tactics. Using our description languages on multiple teams at different skill levels can ensure that our language and data can be applied to the overall volleyball field and not just to specific teams, play strategies, and patterns. Our dataset contains $1,632$ rallies and $12,112$ action features selected from 2019-2021 NCAA Division 1 Big West Conference---including Hawaii vs Long Beach, UCSB vs CSUN, and UCSD vs UCI---, and national team men's volleyball matches---including Japan vs Venezuela, etc. We use team A to represent home teams and B to represent visiting teams. 

\subsection{Dataset Analysis}

Here we perform some simple reality checks of our data against common play strategies, ball handling patterns, and player involvement in today's volleyball games. The analysis is to validate that our data collection statistics adhere to those observed in professional and college level games. It also demonstrates the potential of our data and description for useful tactical analysis, to be discussed in later sections. 

Table~\ref{var_de} analyzes the three variables that volleyball experts believe to have a significant relationship with scoring. We found that the two outside hitters received the most balls, accounting for 50.7\% (bic + outside) of the total amount of sets. On average, each outside hitter received 25.4\% (4\% bic + 21.4\% outside) of the balls from the setter. The opposite and two middle blockers received 26.8 \% ( 6.3\% d-ball + 20.5\% oppo) of the balls and 20.5\% (quick) of the balls from the setter, respectively. According to volleyball experts, these percentages of balls received by attackers at different positions align with the overall tactical trends in volleyball today. Additionally, it is clear that there are heavy preferences for hitting (spiking) the ball (as compared to slower offensive actions) and for jump serving; both are directly in line with current trends in high-level volleyball today. Therefore, our dataset can objectively reflect the current mainstream playing style and tactics of high-level volleyball.

\begin{table}[ht]
\centering
\caption{ Breakdown of the prevalence of different setting locations,\\ offensive actions, and serve type within the full dataset.}
\label{var_de}
\begin{adjustbox}{totalheight=0.17\textheight}
\begin{tabular}{cccccc}
\Xhline{1.5pt}
\makecell{Variable}&\makecell{Label (location or move type)} &\makecell{ Prevalence of each label (\%)} \\
 \hline
 \multirow{6}{*}{\makecell{set\_location}} & outside & \makecell{42.7}\\
  & d-ball & \makecell{6.3}\\
  & oppo & \makecell{20.5}\\ 
  & quick & \makecell{20.5}\\
  & bic &  \makecell{8}\\
  & dump &  \makecell{2}\\
  
  \hline
 \multirow{4}{*}{\makecell{hit\_type}} & hit & \makecell{59.8}\\
 & blocked & \makecell{15.8}\\
 & roll\_shot \& tip \& off\_speed \& dump & \makecell{17.2}\\
 & free\_ball \& overpass & \makecell{7.2}\\
 
 \hline
 \multirow{3.5}{*}{\makecell{serve\_type}} & jump & \makecell{77.1}\\
 & float & \makecell{17.9}\\
 & hybrid & \makecell{5}\\

\Xhline{1.5pt}
\end{tabular}
\end{adjustbox}
\end{table}

\begin{figure}[htbp]
\centerline{\includegraphics[width=7.5cm, height=4.4cm]{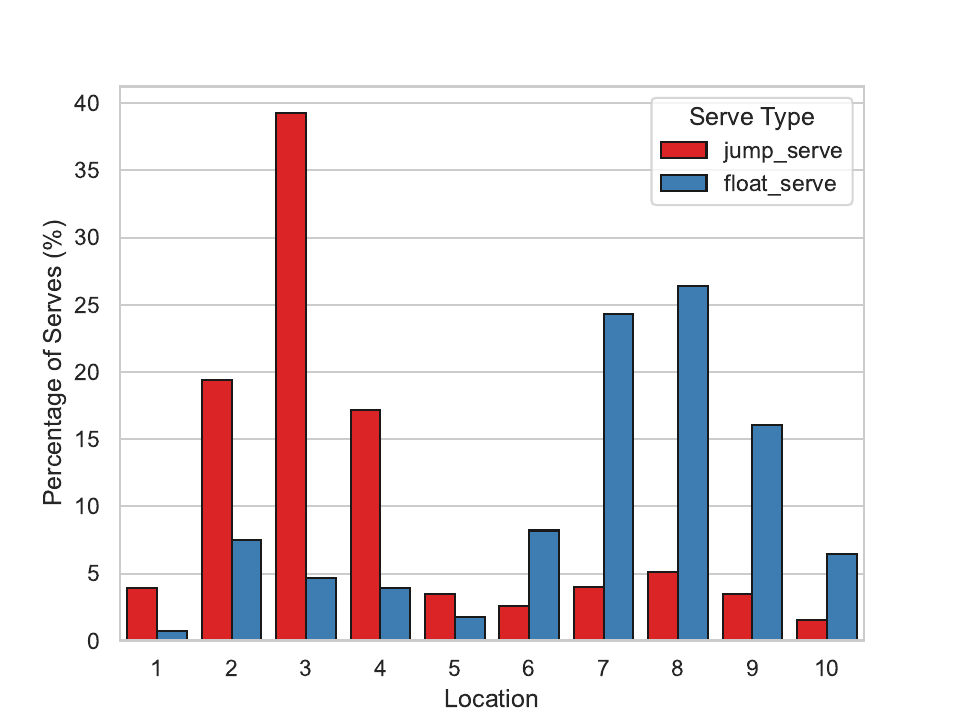}}
\caption{Serve location breakdowns for float serves and jump serves. Notice how jump serves mostly land in the back of the court in areas 2, 3 and 4 while float serves mostly land in the middle of the court in areas 7, 8 and 9.}
\label{fjsld}
\end{figure}

The receiving serve location distribution in Figure~\ref{fjsld} shows the different areas where float serves and jump serves are received. In particular, combined with the court location (Figure~\ref{fig1}), we observe that the receiving points of float serves are mainly in the center of the court, marked as areas 7, 8, and 9. On the other hand, the receiving points of jump serves are mainly in the back of the court, marked as areas 2, 3, and 4. The reason for these regional characteristics is that in high-level volleyball, players prefer to stand in the front of the court to receive a float serve with an overhand pass for more control. When facing the faster jump serve, players choose to receive the serve with an underhand pass at the back of the court to give themselves more time to react to the serve.

\begin{figure}[htbp]
\centerline{\includegraphics[width=8cm, height=4.4cm]{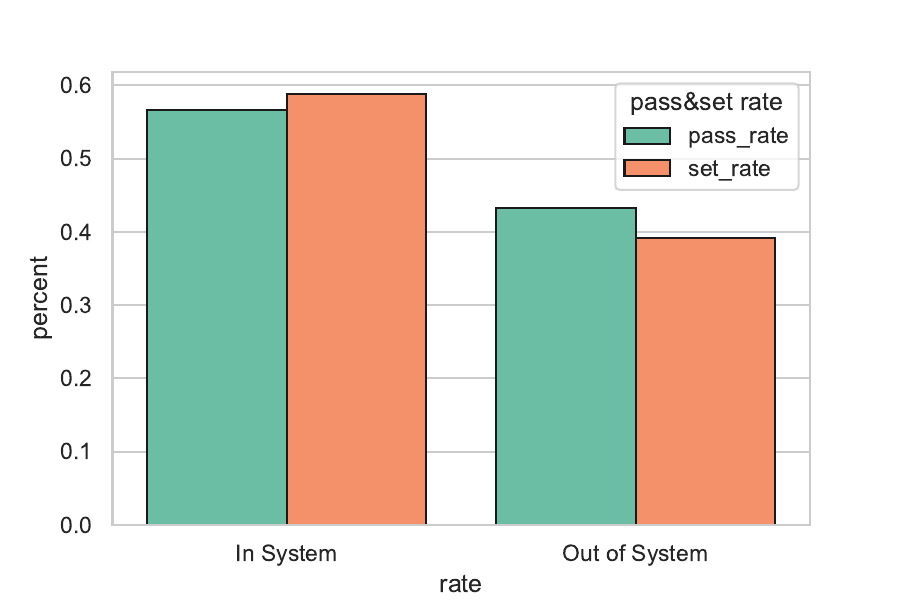}}
\caption{The Rally Winning Probability Breakdown by Round }
\label{sqd}
\end{figure}

Passing and receiving in and out of system is also an important indicator of whether the dataset meets the realities of high-level play. According to volleyball experts, the setter should have the ability to convert out-of-system passes into in-system sets at a high level of play. If this is not the case, the match will not be considered high level. In summary, for a high-level match, the number of in-system sets should be greater than the number of in-system passes, and the number of out-of-system sets should be less than the number of out-of-system passes. Figure~\ref{sqd} follows the definition of serving and receiving patterns for high-level play by volleyball experts. Overall, based on the experts' comments and the above comparison of our dataset with the plays and patterns of the professional games, we can observe that both the professional games and the NCAA games in our dataset are in line with the modern overall trends in high-level volleyball. Including college level matches makes our dataset more diverse and functional, which shows how VREN can be applied and used for analysis outside of the highest competition level.

\section{VREN Tasks}

In this section we illustrate the use of our descriptive language for important tasks in volleyball training and coaching, including predicting rally results, setting types and hitting types, and attacking zones and tactics. The ability to make such predictions accurately is novel and important: As the tasks used to consume hours of reviewing video playbacks can now be performed using our AI-based system much more efficiently. The confirmation and discovery of winning factors in volleyball matches allows trainers and coaches to pay attention to the styles and positions of plays and players' actions. Hence, our descriptive languages and analysis systems can become an invaluable tool for both players and coaches at professional and college levels. 

\subsection{Rally Result Prediction}

Assessment of an opponent's advanced tactics requires coaches to repeatedly review entire game videos, which is  time consuming. In this section, we discuss the new task enabled by VREN, Volleyball Rally Prediction (VRP), to help coaches quickly understand the team's tactical patterns in each rally and provide insights to coaches about their team's winning probability for each round. Moreover, VRP and our dataset allow coaches to simulate a round and see how different tactics may affect rally outcomes. Coaches can simulate a new round scenario by changing labels in VREN. Using the original test data and results as a reference, they can verify whether they have made the right decision by checking whether the predicted winning probability of each round and the final rally predicted outcome can be improved by adopting new tactics.

\subsubsection{Method}
This problem can be framed as follows: Given a set of rounds \{$r^{(n)}$,$y^{(n)}$\}, where $r^{(n)}$ represents a sequence of VREN locations and movements for a single round and $y^{(n)}$ contains the information of which team wins the rally (with \textbf{winning\_reason} and \textbf{losing\_reason} eliminated from $I^{R}$), we can attempt to predict $y^{(n)}$ using $r^{(n)}$ and previous rounds $r^{(k)}$ for $k<n$.

To analyze the efficacy of the data set to predict rally outcomes, we tested 4 different models with a wide range of complexities. First, we used a multi-variate logistic regression \cite{b18} \cite{b28} to perform a simple and very time-efficient linear mapping of our input data to an output probability value. Second, we used a Convolution Neural Network (CNN) \cite{b19} including a 1D Convolution Layer, a hidden Dense Layer with 32 hidden neurons, and an output Dense Layer. Third, we used a Long Short-Term Memory (LSTM) model \cite{b17} including one LSTM layer and an output Dense Layer to handle variable length rallies. Each input we fed into the LSTM model included a fixed number of previous ball rounds across the current rally and previous rallies. Lastly, we used a Transformer Model \cite{b16} --- a powerful architecture that achieves better performance on long sequence tasks --- including 4 Transformer Encoder blocks, a Global Pooling Layer, a Dense Layer with 128 hidden neurons, a Dropout Layer with 40\% dropout, and an output Dense Layer. Each Transformer Encoder block has a Multi-Head Attention Layer with 4 heads and a 25\% dropout, a constricted feed-forward network with two 1D Convolution Layers and a Dropout Layer of 25\% dropout all with only 4 hidden neurons, and normalization after both the Multi-Head Attention Layer and the feed-forward network.

\subsubsection{Experimental Setup}
\begin{itemize}
\item \textbf{Implementation Details.} We trained the majority of our models on our VREN dataset with roughly 80\% of the total sequences. The remaining 3 matches not used in training were reserved as one validation set and two testing sets (one professional level game and one college level game) with 7\% and 13\% of the total sequences for evaluating the performance of models. 

\item \textbf{Evaluation Metrics.} We adopted four statistical evaluation metrics: Binary Accuracy (accuracy of two class classification), AUC \cite{b29} (area under the ROC curve, a measure of probabilistic prediction performance), Brier Score \cite{b30} (mean squared error for probabilistic values), and Mean Absolute error \cite{b29} (linear distance of prediction from true value), to evaluate the performance of our VRP models. 
\end{itemize}

\subsubsection{Results and Analysis}
Table~\ref{model_res} shows the results of the four models on both professional and college level games. For both competition levels, the performance of the Transformer model is the best among all four models. At the college level, the accuracy of the Transformer is 5.32\% higher than that of the CNN, which ranks second in accuracy. At the professional level, the accuracy of the Transformer is 7.27\% higher than that of the logistic regression, ranked second in accuracy. In general, all four models predicted better results at the professional level than at the collegiate level. After consulting volleyball experts, we believe that the reason for this result is that professional teams have better discipline and more advanced and mature skills than college teams, leading to noticeably more deterministic, and thus predictable, outcomes. Thus the difference in the predicted results at different levels is as expected. These results could likely be improved with new encoding methods and more specialized models, but we leave this for future study.
\begin{table}[ht]
\centering
\caption{performance of each model on different metrics for the\\ VRP on a college level game \& a professional game. \\LG refers to Logistic Regression, TR refers to Transformer}
\label{model_res}
\begin{adjustbox}{totalheight=0.09\textheight}
\begin{tabular}{cccccc}
\Xhline{1.5pt}
\makecell{Level of game}&\makecell{Model} &\makecell{Binary Accuracy(\%)}&\makecell{AUC}&\makecell{Brier Score}&\makecell{Mean Absolute Error} \\
 \hline
 \multirow{4}{*}{\makecell{college}} 
  & LG & 66.56 & 0.66 & 0.33 & 0.33\\
  & CNN & 69.06 & 0.75&0.20 &0.40\\
  & LSTM & 65.91& 0.75&0.21 &0.41\\
  & TR & 74.38& 0.82&0.18 &0.34\\
  \hline
 \multirow{4}{*}{\makecell{professional}}& LG & 72.73 & 0.72 & 0.28 & 0.28\\
  & CNN & 71.59 & 0.76& 0.20 &0.39\\
  & LSTM & 70.06& 0.75& 0.20 &0.40\\
  & TR & 80.00& 0.85& 0.16 &0.32\\
 
\Xhline{1.5pt}
\end{tabular}
\end{adjustbox}
\end{table}

\begin{figure}[htbp]
\centerline{\includegraphics[width=8cm, height=4.6cm]{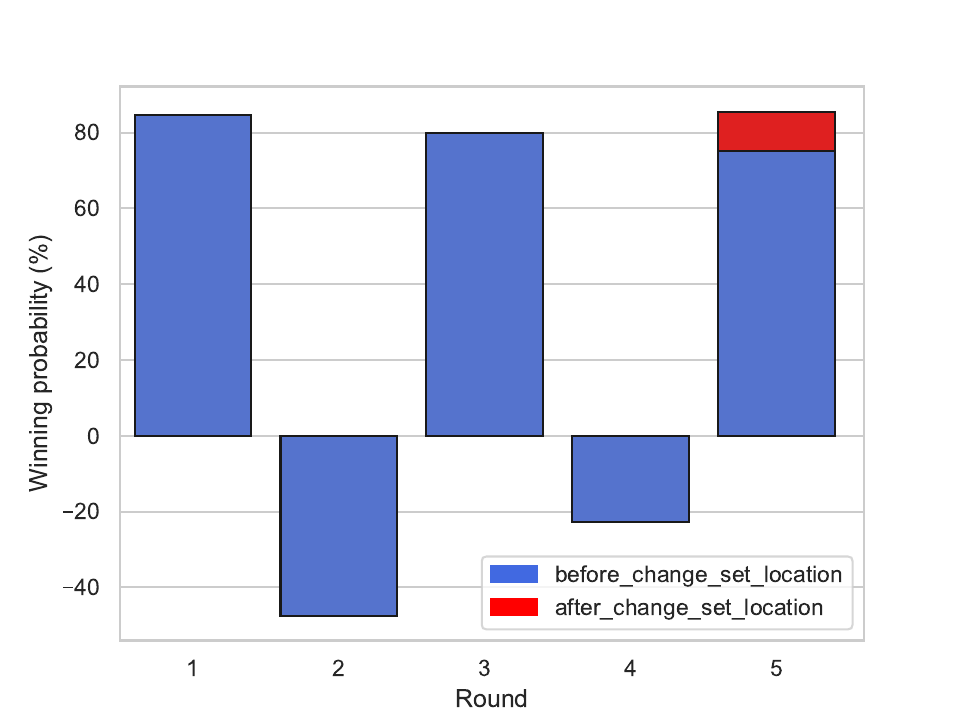}}
\caption{Winning probability of the rally breakdown by round}
\label{RWP}
\end{figure}

 Using these models, we can also test how changes in tactics may lead to improved winning chances as predicted by our trained VRP model. One such scenario we analyze is detailed in figure~\ref{RWP}. In this figure, the blue bar charts show the probability of each team winning the point at the end of each volleyball round in the rally using the original data in our dataset. The red bar represents the increase in the probability of winning after changing a single tactical variable in the last round. In this example, we changed the last set location--the location where the setter sets the ball to--from a d-ball to a quick based on the volleyball expert's advice. As a result, increased probability of the offensive team winning by about 10\%. In particular, rounds 1, 3, and 5 mean that the ball is on team A's side, and rounds 2 and 4 mean that the ball is on team B's side. This example illustrates one way in which our VRP can be used in practice; coaches can adjust their tactics and drills based on the information they want to change in the set, combined with the probability of winning the set.

\subsection{Setting Type and Hitting Type Prediction} 
According to volleyball experts, the types of set and hit are almost always the two most important factors that a team needs to judge and predict when defending. For the middle blockers, judging the opponent's setting position in advance can provide them with more time to move, so that they can cooperate with other blockers to form a more effective blocking screen. Furthermore, knowing the opponent's attacking style in advance can enable them to choose the most suitable blocking style and technique. The back row defenders usually choose different defensive skills and formations when facing different attackers and different hitting types. Making a judgment in advance can help them set up corresponding defensive tactics and effectively improve the success rate of their defensive strategy. Our goal for this task is to assist the team in defensive judgment training and allow them to develop better situational awareness and judgment in the game.

\subsubsection{Method}
Predicting the type of sets can be framed as follows: Given a set of rallies \{$r^{(n)}$,$y^{(n)}$\}, $r^{(n)}$ represents a sequence of VREN locations and movements without setting type information and $y^{(n)}$ only contains the \textbf{type of sets} information. Instead of containing the winning team information, $y^{(n)}$ includes only nine types of sets (quick, outside, oppo, bic, d-ball, dump, overpass, blank/no set, and blocked). Therefore, using all the current ball round's information leading up to a given set in a rally, we attempt to predict where that setter will set the ball.

Similarly, hitting type can be framed as follows: Given a set of rallies \{$r^{(n)}$,$y^{(n)}$\}, $r^{(n)}$ represents a sequence of VREN locations and movements without \textbf{hitting type} information and $y^{(n)}$ only contains the type of sets information. Instead of winning team information, $y^{(n)}$ only includes nine hitting types: hit, off\_speed, roll\_shot, tip, free\_ball, dump, overpass, blocked, blank). Therefore, using all the current ball round's information leading up to a given hit in a rally, we attempt to predict what attacking style that hitter will use.

We used the same Transformer architecture as Task One; we modified its usage for classification instead of regression to set up the Transformer model to predict $y^{(n)}$ based on $r^{(n)}$. 

\subsubsection{Experimental Setup}
\begin{itemize}
\item \textbf{Implementation Details.} We trained a Transformer model on our VREN dataset using approximately 80\% of the total sequences as we did in task 1. The remaining three matches were respectively used as one validation set and two testing sets (one professional level game and one college level game) with 7\% and 13\% of the total sequences for evaluating the performance of models. 

\item \textbf{Evaluation Metrics.} Our Transformer model was evaluated using categorical accuracy.  

\end{itemize}

\subsubsection{Results and Analysis}
\begin{table}[htbp]
\caption{Categorical Accuracy for setting location prediction \\and hitting type prediction in both professional and college level games}
\label{sha}
\begin{center}
\begin{adjustbox}{totalheight=0.07\textheight}
\begin{tabular}{cccc}
\hline
\textbf{Predicted Value\quad } & \textbf{\textit{Competition level}}& \textbf{\textit{Categorical accuracy\%}}\\
\hline
\multirow{2}{*}{hitting type   \quad } & College level & 71.28\\
& Professional level & 73.63\\
\hline
\multirow{2}{*}{setting location   \quad  } & College level & 54.65\\
& Professional level & 51.65\\
\hline
\end{tabular}
\end{adjustbox}
\label{tab1}
\end{center}
\end{table}
Table ~\ref{sha} shows that the Transformer's hitting type prediction accuracy is 71.28\% at the NCAA competition level and 73.63\% at the professional level. Similar to the predicted result of Task 1, the Transformer has a better performance at the professional level than at the college level. The difference between the Transformer's performance for the two levels of hitting type is not significant, which indicates that there is not much difference between college level players and professional players regarding offensive options. However, in predicting the setting type, the Transformer has 54.65\% categorical accuracy for the college level compared to the 51.65\% accuracy for the professional level. This is the first time in all of the prediction tasks that the college level prediction results is better than the professional level prediction results. According to experts, the prediction result for the setting type occurs because it is highly dependent on the skill level of the setter. The setting is more regular and relatively easy to predict at the college level. However, in professional games, the setter will often make some unconventional sets to break opponent defenders' habits, which makes it more difficult for our model to predict.

Note that these prediction tasks are crucial for volleyball preparation and training. 
Predicting the setting type is one of the most important tasks for coaches, as using optimal defensive strategies for a given set type can greatly improve a team's chance of winning. If the predictions are successful, it will help coaches and players prepare more optimally for the game. A coach can use the prediction results to determine the overall tendencies of the opponent's setter in different situations and make targeted defensive arrangements, which will ultimately improve the team's chances of winning.
In this task, we propose a prediction baseline for the setting type and hitting type. Due to the importance of these variables in the game of volleyball, we wish to investigate more complex models and embeddings to boost performance and bring a novel approach to volleyball tactical analysis.

\subsection{Volleyball Tactics and Attacking Zone Statistics} 

Effective and timely statistics in volleyball have a significant impact on the coach's tactical choices in the game. Today's volleyball statistics are marked by manual observation of the game. In this case, the information that volleyball statistics can provide, such as the number of team and individual errors, points, blocks, serves, etc., is limited. Furthermore, existing volleyball statistics are unable to provide more detailed decision support for coaches. By introducing VREN and our dataset, volleyball statistics can be furnished in realtime in a more detailed and informative manner. For example, we can clearly reflect the area where the ball falls and the hitting line as well as the overall offensive strategy being used through the position encoding information in VREN. /// We can judge the tactics used through the position of the attacker and the relative position between the attacker and the setter and also provide the coach with relevant information about the opposing team's attacking style choice by referring to the hitting type. With our dataset, we are able to provide not only all of the existing volleyball statistics, but also more detailed and advanced technical statistics generated by a quick consultation of VREN. The result is a new level of support for on-court decisions and pre-game analysis for coaches that is difficult, if not impossible, to achieve with existing technical statistics. 

\subsubsection{Methods}

\begin{figure}[htbp]
\centerline{\includegraphics[width=3cm,height=3cm]{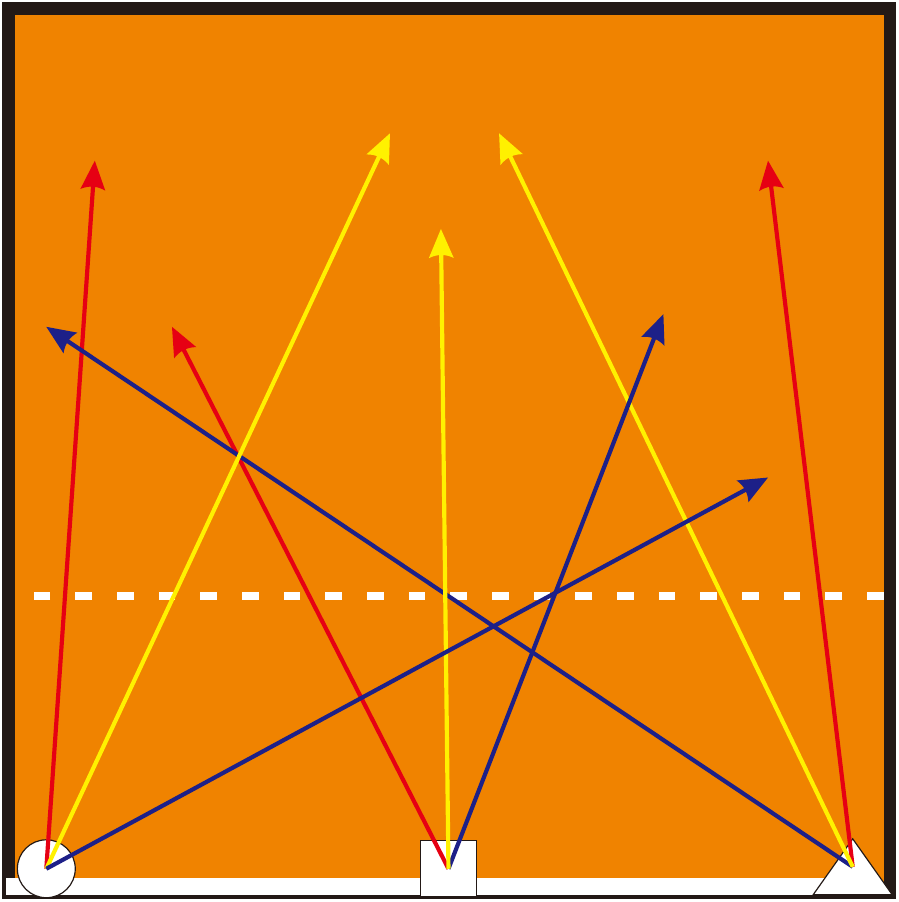}}
\caption{ The three graphics at the bottom represent the attackers. The circle represents the outside hitter, the square represents the middle blocker, and the triangle represents the opposite. The colored lines represent hitting direction. Red represents line, blue represents angle, and yellow represents seam.}
\label{hlc}
\end{figure}

We created a Python script to be used in conjunction with our dataset to provide detailed volleyball statistical information proposed by experts. Those statistics include the proportion of attackers hitting different general locations, the proportion of setters setting the ball to different positions when the pass is in and out of the system respectively, the proportion of team attacking tactics applied, etc. Specifically, general attacking location information is not shown in our dataset but can be framed as follows: Given these sets for our grid location system: s1: \{1, 2, 6, 7\}, s2: \{4, 5, 9, 10\}, s3: \{3, 8\}, s4: \{11, 12\} s5: \{14, 15\}. We use x to represent the number of balls hit straight along the sideline (typically called line), y to represent the number of balls hit sharply across the court (typically called angle), z to represent the number of balls hit toward the middle of the court (typically called seam), and b represents the receiving location. For better understanding of these terms, \autoref{hlc}. provides a hitting location schematic diagram. When the outside hitter hits the ball, if ${b \in s1}$, then we increment x, if ${b \in s2\cup s5}$ we increment y, and if $b \in s3$ we increment z; when the middle blocker hits the ball or the back row outside hitter hits the bic, if ${b \in s1 \cup s4}$ then we increment x, if ${b \in s2 \cup s5}$ we increment y, and if $b \in s3$ we increment z; when the opposite hits the ball, if ${b \in s2}$ then we increment x, if ${b \in s1 \cup s4}$ we increment y, and if $b \in s3$ we increment z. The proportion of general attacking locations can be calculated as follow:
\begin{itemize}
\item percent hit line: $\frac{x}{(x+y+z)} * 100\%$
\item percent hit angle: $\frac{y}{(x+y+z)} * 100\%$
\item percent hit seam: $\frac{z}{(x+y+z)} * 100\%$
\end{itemize}

\subsubsection{Experimental Setup}
\begin{itemize}
\item \textbf{Implementation Details.} We applied our python script to one professional volleyball match from our dataset. We first used our set location variable to get the ratio of in-system vs. out-of-system sets. Moreover, we calculated the proportion of offensive tactics used in- vs. out-of-system sets, as well as the proportion of offensive motions chosen by the players in the corresponding tactics and the proportion of the players' offensive lines.


\item \textbf{Evaluation Metrics.} Our statistics are reviewed by volleyball experts.
\end{itemize}

\subsubsection{Results and Analysis}

Table ~\ref{tab1} shows the results of our volleyball statistics. According to volleyball experts, the defensive tactics when the other team passes in system are entirely different from the defensive tactics for out-of-system passes. Therefore, in our volleyball statistics, we separate the in- and out-of-system situations. In system meaning the pass or set are nearly perfect and all offensive options are available, and out of system meaning the pass or set are not ideal and not all offensive options are available. For our experimental match, Team A had 69.81\% in-system sets and 30.19\% out-of-system sets. The hitting (spiking) rate was high in all positions and above 85\% when sets were in system. This data shows that when team A's set is in system, team B's defenders and blockers have to prepare more for the opponent's spike. Analysis of general hitting locations of different positions in these statistics can let the defenders know where a spike will go with a high probability under different circumstances so they can make corresponding defensive arrangements in advance. For example, when team A's middle blockers use the tactic of a "thirty-one" (a set to the middle with a gap from the setter, usually to zone 14), 71.29\% of them hit the ball in a straight line. In this case, team B's defense and blocking against team A's middle blockers should mainly cover the straight line. 

When the pass is out of system, the statistic is completely different from when the set is in system. 0\% quick and 0\% "thirty-one" means that Team A's middle blockers do not have a single attack when the set is out of system. Thus team B can ignore the attack from team A's middle blockers when the set is out of system and focus on the other attackers. In addition, the hitting rate and the percentage of line selection vary a lot compared to the data when the set is in system. Unlike in-system sets---where the hitting rate is above 85\% for all setting options---, hitting rates for outside and oppo are only 62.5\% and 50\%, respectively in out-of-system scenarios. Furthermore, the percentage of hitting the middle of the court is also 0\%, which means that defenders should cover other locations when there is an out-of-system set. Moreover, outside hitters are more likely to choose to hit a straight line, and the opposite is more likely to hit a diagonal line (angle).

Our statistics differ from existing technical statistics in that they are more detailed and provide coaches with better on-court decision aids. Our volleyball experts believe that, if our volleyball statistics can be entered and analyzed in real-time by distinguishing different players, locations, and actions automatically, it will make a large impact on the game of volleyball by allowing coaches to make more informed tactical decisions on the fly. We will leave incorporating computer vision strategies to enable real-time data input and statistics analysis by expanding on these naive methods for future study.


\begin{table*}[htbp]
\centering
\caption{More detailed statistics include a breakdown of set ratings and locations, an analysis of hitting locations, and an evaluation of attacking move distributions.}
\begin{center}
\begin{adjustbox}{totalheight=0.14\textheight}
\begin{tabular}{ccccccccc}
\toprule
set rating            & \begin{tabular}[c]{@{}c@{}}overall  share of sets\end{tabular} & set location & \begin{tabular}[c]{@{}c@{}}Breakdown by  set location (\%)\end{tabular} & \begin{tabular}[c]{@{}c@{}}percentage of  spike (hit) (\%)\end{tabular} & \begin{tabular}[c]{@{}c@{}}percentage of  junk (roll \\shot, tip,  \& off speed) (\%)\end{tabular} & \begin{tabular}[c]{@{}c@{}}percentage \\ hit line (\%)\end{tabular} & \begin{tabular}[c]{@{}c@{}}percentage \\ hit angle (\%)\end{tabular} & \begin{tabular}[c]{@{}c@{}}percenage \\ hit seam (\%)\end{tabular} \\ \midrule
\multirow{6}{*}{in system}  & \multirow{6}{*}{69.81\%}                                  & outside      & 23.68\%                                                                   & 88.89\%                                                          & 11.11\%                                                                                   & 25.00\%                                                                & 37.50\%                                                                 & 37.50\%                                                                           \\
                            &                                                           & bic          & 18.42\%                                                                   & 100.00\%                                                         & 0.00\%                                                                                    & 57.14\%                                                                & 14.29\%                                                                 & 28.57\%                                                                           \\
                            &                                                           & oppo         & 18.92\%                                                                   & 85.71\%                                                          & 14.29\%                                                                                   & 50.00\%                                                                & 16.67\%                                                                 & 33.33\%                                                                           \\
                            &                                                           & d-ball       & 8.11\%                                                                    & 100.00\%                                                         & 0.00\%                                                                                    & 25.00\%                                                                & 50.00\%                                                                 & 25.00\%                                                                           \\
                            &                                                           & thirty  one  & 18.92\%                                                                   & 100.00\%                                                         & 0.00\%                                                                                    & 71.43\%                                                                & 14.29\%                                                                 & 14.29\%                                                                           \\
                            &                                                           & quick        & 10.81\%                                                                   & 100.00\%                                                         & 0.00\%                                                                                    & 25.00\%                                                                & 75.00\%                                                                 & 0.00\%                                                                            \\ \midrule
\multirow{6}{*}{out system} & \multirow{6}{*}{30.19\%}                                  & outside      & 50.00\%                                                                   & 62.50\%                                                          & 37.50\%                                                                                   & 80.00\%                                                                & 20.00\%                                                                 & 0.00\%                                                                            \\
                            &                                                           & bic          & 12.50\%                                                                   & 100.00\%                                                         & 0.00\%                                                                                    & 50.00\%                                                                & 50.00\%                                                                 & 0.00\%                                                                            \\
                            &                                                           & oppo         & 25.00\%                                                                   & 50.00\%                                                          & 50.00\%                                                                                   & 50.00\%                                                                & 50.00\%                                                                 & 0.00\%                                                                            \\
                            &                                                           & d-ball       & 12.50\%                                                                   & 100.00\%                                                         & 0.00\%                                                                                    & 0.00\%                                                                 & 100.00\%                                                                & 0.00\%                                                                            \\
                            &                                                           & thirty  one  & 0.00\%                                                                    & NA                                                               & NA                                                                                        & NA                                                                     & NA                                                                      & NA                                                                                \\
                            &                                                           & quick        & 0.00\%                                                                    & NA                                                               & NA                                                                                        & NA                                                                     & NA                                                                      & NA                                                                                \\ \bottomrule
\end{tabular}
\end{adjustbox}
\label{tab1}
\end{center}
\end{table*}

\section{Conclusions and Future Work}
In this paper, we introduce a new language, VREN, to describe volleyball games in a formatted way. In addition to the language, we captured information using the language and introduced a new high-quality dataset for high-level volleyball games. Experts believe that the proposed dataset has the potential to bring an entirely new level of player development and tactical analysis to volleyball. Based on these experts' suggestions, we propose three volleyball tasks that can assist coaches in improving their decision-making and tactics: volleyball statistics, volleyball round prediction, and setting/hitting type prediction. With our dataset, we can improve upon existing statistics by including more detail to yield much more tactical information. We propose deep learning models for volleyball rally winner, setting type, and hitting type predictions and use the results of our models as a baseline for new models in the future. In conclusion, this paper bridges the gap between the field of volleyball and computer science. Volleyball data analysts can use our language directly to retrieve valid and useful information from game data, and hence, reduce their burden of designing data formats and reviewing game videos. Moreover, players and coaches can improve their tactics while finding the weaknesses of their opponents. Furthermore, our data representation can contribute to other rally-type sports research fields such as beach volleyball and doubles tennis with quick revisions of some small features of our language representation.

In our future research, we plan to expand our dataset by adding more data, propose more sophisticated models to improve accuracy, and incorporate computer vision materials to automatically label inputted video data according to our VREN representation. Through the close integration of volleyball and computer science, we hope that our models, statistics, language, and dataset will eventually help players and coaches establish a different way of thinking about volleyball tactics and bring a new perspective to volleyball training and tactical development. Additional future work we would like to explore includes using computer vision to predict the landing location of a serve using a server's posture. This information would assist a passer with judging the optimal passing position and tactic in advance to improve the success rate of the pass.

\vspace{12pt}
\end{document}